# Opinion paper. Generative AI and the future of scientometrics

**Benedetto Lepori**, Università della Svizzera italiana, via Buffi 13, 6904 Lugano, Switzerland; blepori@usi.ch, https://orcid.org/0000-0002-4178-4687.

**Jens Peter Andersen**, University of Aarhus, Bartholins Allé 7, DK-8000 Aarhus C, Denmark; jpa@ps.au.dk, https://orcid.org/0000-0003-2444-6210.

**Karsten Donnay**, University of Zurich, Affolternstrasse 56, 8050 Zurich, Switzerland; donnay@ipz.uzh.ch, https://orcid.org/0000-0002-9080-6539

**Abstract**

In this paper, we contribute to the debate on generative artificial intelligence (GenAI) in scientometrics. We argue that moving from a trial-and-error approach to an explainable and actionable use requires a principled understanding of strengths and weaknesses of GenAI as compared with other techniques and with human judgment. To this end, we introduce a conceptual framework based on the distinction between the semantic dimensions of texts, i.e. the meanings attributed to words, and their pragmatic dimension, i.e. their embedding within communicative situations. We leverage this framework to interpret the results of applications of GenAI in scientometrics and to provide guidance to users. Specifically, we conclude that key parameters to be considered are the nature of the task, the level of granularity of the analysis and whether the goal was descriptive, inferential or evaluative. These parameters lead to different strategies for using GenAI and human-machine integration. Finally, we suggest that, by generating large amounts of scientific language, GenAI might affect textual characteristics used to measure science, such as authors, words, and references. We argue that careful empirical work and theoretical reflection will be essential to remain capable of interpreting the evolving patterns of knowledge production in the age of AI.

Keywords: scientometrics; generative AI; authorship; chatGPT; large language models; scientific language; semantics; pragmatics.

# 1 Introduction

Generative Artificial Intelligence (GenAI) tools are being increasingly adopted in scientific research (see Rahman et al., 2023 Chen, Q. et al., 2025). The field of scientometrics, broadly defined as the quantitative study of science and technology (Hood & Wilson, 2001; Bar-Ilan, 2009), is no exception. In the last five years, a flurry of studies has been published testing the use of GenAI in scientometrics (for a recent review see Zhang, Yi et al., 2025). This includes the identification of emerging research topics (Lu et al., 2024), labelling research topics (Kozlowski et al., 2024; Murray et al., 2025), predicting citation counts (de Winter, 2024), and identifying prominent scholars (Sandnes, 2024). Other emerging applications of GenAI are the evaluation of research quality (Thelwall, 2024), the categorization of research papers based on metadata (van Eck and Waltman 2024), the identification of disruptive research (Bornmann et al., 2024) and novelty evaluation of scientific articles (Wu et al., 2025).

Most of the existing works to date adopt an empirical approach where outputs produced by GenAI tools are compared with existing techniques and/or human judgment. Much fewer studies have engaged with the technical aspects put forth in the GenAI evaluation literature (Schneider, 2024) to improve the reliability of results (Kozlowski et al., 2024). Existing studies also rarely apply best practices for systematically evaluating and optimizing GenAI performance such as validation (Murray et al., 2025), prompting (Cantone et al., 2025) and training (Zhu, M. et al., 2025).

There is also no clear consensus yet on the usefulness of GenAI in scientometrics. Some authors have highlighted its potential for quantitative science studies, such as access to data without the need for specialized skills (Thelwall, 2024), ease of interaction (Lepori et al., 2025), and the ability to combine information and to develop categorizations (van Eck & Waltman, 2024). Others, in contrast, have argued that the quality of results falls short of the current standards in terms of stability, reliability and quality (Sandnes, 2024; Thelwall, 2025c).

The goal of our paper is to advance the debate on two central questions related to the applicability of GenAI approaches for the field of scientometrics.

*First, can we derive criteria that allow us to anticipate the performance of GenAI for typical applications in scientometrics research? More specifically, can we identify characteristics of the task that enable us to evaluate the potential and limitations of GenAI, and to devise robust adoption strategies?*

We address this question by building on the linguistic foundations of GenAI as computational techniques. Specifically, in section 2, we propose a framework that interrogates the performance of GenAI based on the core dimensions of written texts in their communicative function (Morris, 1938), i.e. their format (syntax), the meanings (semantics) and the relationships to the interpreter (pragmatics; Sobieszek & Price, 2022; Piantadosi, 2023; Loru et al., 2025).

In section 3, we build on this framework to interpret recent experiments of GenAI use in scientometrics by distinguishing between tasks mostly related to understanding meanings in texts (classification, topic labelling, etc.) and tasks mostly related to practical scientometric activities (identifying prominent scholars; research evaluation; etc.). We suggest that the underlying architecture of GenAI leads to better and more consistent performance for the former than for the latter. We also propose strategies to improve quality and reliability by leveraging techniques such as calibration, validation, prompting, or human-machine integration. Our framework, therefore, provides guidance to evaluate GenAI performance as related to the nature of the task and to the structural features of the underlying linguistic models.

*Second, looking into the future, we discuss implications of the (automated) generation of large amounts of scientific language in academic publications for the field of scientometrics, a research area that relies on the quantification of textual content to indirectly observe scientific practices (*Wouters, 1999b*).*

We address this question in section 4 providing an exploratory discussion of the potential impact of GenAI on core scientometrics measures, such as authors (Kousha, 2024; Gorraiz, 2025), scientific productivity (Kusumegi et al., 2025), vocabulary (Kobak et al., 2024; Bao et al., 2025), and references (Eger et al., 2025). For each of these dimensions, we advance hypotheses on the extent to which GenAI use might require scholars to redefine some of the commonly used measures in scientometrics and their substantive interpretation. We further discuss whether we can still compare post-GenAI to pre-GenAI papers and we suggest empirical analyses that might allow quantifying these effects.

Finally, in section 5, we suggest three key parameters that scholars should consider in designing applications of GenAI in scientometrics, and in interpreting their results, i.e. the linguistic nature of the task, the level of granularity of the analysis, and whether its goal was descriptive, inferential or evaluative. And, beyond the use of GenAI as a tool, we argue for a new theoretical and empirical research program to understand how GenAI is affecting established indicators in our field and the way they represent knowledge production.

# 2 Generative AI and its linguistic foundations

## 2.1 How GenAI works

GenAI is a set of computational techniques that generate new content (textual, images, audio) from training data (Feuerriegel et al., 2024). This generative character distinguishes GenAI from discriminative techniques that aim at dividing entities into groups (such as topic modelling; Ng & Jordan, 2001). At the most fundamental level, GenAI tools return the most probable sequence of tokens, based on input prompts. These prompts are a given sequence of input tokens, such as a question posed by users in a conversational interface.

In today’s GenAI architectures, the generative task is performed by (multimodal) Large Language Models (LLMs), a broad class of models most often based on the transformer architecture introduced by Google in 2017 (Vaswani et al., 2017) and pre-trained on very large amounts of data (Kalyan, 2024). Transformer models analyse all parts of a text simultaneously and learn patterns in how tokens co-occur and relate to one another across many examples of language use. The models’ defining feature is the so-called attention mechanism, which allows each part of a text to be interpreted in relation to all other parts. Input text is treated as a fixed and stable unit, what changes is how strongly different tokens are weighted relative to one another (Vaswani et al., 2017).

In the transformer architecture, meaning thus emerges from the ability of the models to capture (complex) network of relations among tokens within a corpus. This enables the models to handle multiple textual features at the same time: regularities in the structure of token sequences corresponds to language syntax whereas patterns of association between token sequences enable semantic interpretation (Chang, T. A. & Bergen, 2024) and pragmatics are captured through long-range dependencies between token sequences (Ma et al., 2025).

The most well-known LLMs are the GPT family developed by OpenAI (Banik et al., 2024). It is important to distinguish between closed source LLMs, such as GPT and GlaM by Google, and more open alternatives such as Meta's LlaMA (Kalyan, 2024) or the DeepSeek family (Guo, D. et al., 2025). The latter are not fully "open source" but rather "open weights", i.e., the model, including the weights, is publicly available though not the data used for training.

The improvement in LLM performance in recent years largely owes to the increase in the number of parameters and the amount of training data, allowing them to achieve good performance even in tasks for which they were not specifically pre-trained. In addition, models are then typically optimized for specific tasks or purposes. Models used for chatbot interfaces are *aligned* based on extensive Reinforcement Learning from Human Feedback (RLHF) to fine-tune their answers, for example in a sequence of prompts (Feuerriegel et al., 2024). Pre-trained LLMs are also often *fine-tuned* to a specific task or domain by continuing training on task-relevant labelled data (Howard & Ruder, 2018). And model *calibration* evaluates correctness and reliability compared to known reference data (Desai & Durrett, 2020). Alignment, fine-tuning and calibration require *humans-in-the-loop*, i.e., annotators that label model outputs as correct or incorrect, or provide reference data and corrective feedback.

## 2.2 A linguistic framework for understanding GenAI performance

Despite their technical novelty, GenAI tools are built on the long-standing assumption that, by observing statistical patterns of words co-occurrences within a corpus, it is possible to draw inference on the meanings associated with these expressions (Harris, 1954; Sahlgren, 2008).

Effectively, LLMs build a statistical model of language as observed in the corpus on which they were trained (Piantadosi, 2023). The modelling of long-range word relationships (Sobieszek & Price, 2022) allows to produce not only well-formed, but also semantically plausible sentences (Tao et al., 2024). Moreover, the combination of a very large number of parameters and of the size of the pre-training corpus, allows LLMs to become, at least partially, context sensitive, for example to capture how the meanings of words vary depending on the context (Ortega-Martín et al., 2023). A key finding in evaluating LLMs is that performance strongly varies by the nature of tasks (Kocoń et al., 2023; Chen, H., 2023; Chang, Y. et al., 2024).

Since GenAI is mostly used for generating new textual content, we propose a conceptual framework that enables to interpret its performance based on the intended function of the output. Specifically, linguists distinguish between three dimensions (Morris, 1938), i.e. *syntax*, the extent a text is well-formed and respects grammatical rules, *semantics*, i.e. the meanings attributed to words or expressions, and *pragmatics*, i.e. the relationships of a text to its context of use (Grice, 1975; Table 1).

LLMs have excellent command of syntax, usually better than most native speakers (Mahowald et al., 2024). Unlike rule-based text generation, their architecture allows to reproduce language variation in different contexts incl. grammar exceptions, idiosyncrasies etc. This command of syntax is a fundamental requirement for tasks where LLMs produce textual output for further use, like composing a letter, proof-reading a text etc. (Fredrick &

Craven, 2025; Zeleke et al., 2025). But this requirement is less relevant for applications in scientometrics, where the focus is on extracting information from existing content.

Semantics involves tasks where LLMs are required to 'understand' the meaning of a word or of a sentence assuming that the relevant information is available in the text – including for example classification and labelling tasks, coding texts etc. The semantic performance of LLMs has constantly improved with increasing numbers of parameters and the amount of training data (Piantadosi, 2023; Sobieszek & Price, 2022). Thanks to its sensitiveness to context, GenAI also shows a better ability to deal with polysemy, i.e. words having different meanings in different contexts, than traditional NLP methods (Piantadosi, 2023). However, despite high statistical correlations, individual interpretation of meanings might depart from the human understanding, so-called 'hallucinations' or instances of semantic leakage (Gonen et al., 2024). Accordingly, careful calibration and post-hoc validation are needed. Prompting techniques such as chain-of-thought have proven effective to provide more consistent output (Wang et al., 2022).

Finally, pragmatics involves tasks where information on the context of usage is required to infer textual meanings (Grice, 1975; Sperber & Wilson, 1986). The pragmatic dimension comes into play when GenAI is tasked to interpret texts that are closely associated with communicative situations, such as dialogues, or when the generated texts are intended to be put into action, for example when writing a review (Zhou et al., 2024). LLMs have some pragmatic competence, for example they can identify whether a sentence includes humour or irony (Hu et al., 2023). However, they display more variation between tasks and situations (Sravanthi et al., 2024 Ma et al., 2025). GenAI can identify broad patterns – such as typical situations being associated with humour – and its judgments tend to be statistically correlated with human judgment. However, studies shows that it is less adaptive to individual circumstances than humans, which rely on simpler, but more flexible and context-responsive heuristics (Guo, B. et al., 2023; Loru et al., 2025).

Prompt engineering is generally considered an approach suited to improve the pragmatic validity of output generated by GenAI (Ekin, 2023; Bandi & Zeng, 2024). However, it is not always feasible to provide contextual information describing a specific communicative situation. Moreover, in current LLMs, information from prompts is only used to refine results, while the model parameters are set in the pre-training stage. LLMs are in that respect less context-sensitive than humans, which rely on simpler but more flexible heuristics (Mahowald et al., 2024). Accordingly, it has been suggested that human-machine integration would be a more promising avenue for improving GenAI performance in pragmatic tasks (Zhu, M. et al., 2025).

As a summary, our discussion highlights three key aspects to be considered in explaining GenAI performance for scientometric tasks (see Table 1). First, the intended use of GenAI generated output plays a central role for quality – for instance, GenAI performs systematically better for inferring meanings from an existing corpus than for generating human-like judgements. Second, GenAI performs well in providing statistically robust results when variation occurs between broad contexts (such as scientific fields) that can be characterized through prompting, while results are less reliable when dealing with individual occurrences and where more flexible heuristics might be relevant for judgement (such as when reviewing an individual paper). Third, the way GenAI is integrated with human expertise is fundamental for proper usage, combining human heuristics for communicative situations with the reliance of GenAI on broad statistical patterns.

*Table 1. Textual functions and GenAI*

| | **Syntax** | **Semantics** | **Pragmatics** |
|---|---|---|---|
| Definition | Practical rules for text formation | Meaning of words as can be inferred from a corpus | Relation of a text to a specific communicative situation |
| Dependence on context | Low (e.g. American vs. British English) | Medium (general contextual dimensions, like the field, the generic communicative situation) | High (individual features of a communication context) |
| Examples of tasks | Proof reading texts | Labelling, classifying, summarizing general knowledge | Generating speeches, letters, etc. Interpreting dialogues. |
| LLM performance | Excellent | Very good, some hallucinations | Mixed, and very variable |
| Prompting | May be useful | Very useful to characterize the communicative situation | Limited efficacy as situations often very specific |

# 3 GenAI as a scientometric tool

In revisiting recent applications of GenAI in scientometrics, we propose leveraging the conceptual framework introduced in the previous section to distinguish between tasks according to the dominant textual function they require, namely semantic inference and pragmatic judgment.

We focus on applications that fall within the core areas of scientometrics, i.e. tasks aimed at extracting, structuring or interpreting information from scientific texts and metadata. Uses of GenAI for writing, rewriting, or otherwise producing or reviewing (including peer review) scientific texts are discussed in section 4, to the extent that they affect the empirical basis of scientometric analysis.

## 3.1 Inferring meaning from text

A first group of applications relies primarily on GenAI’s ability to infer meaning from textual corpora. These tasks require sensitivity to context and polysemy, but do not require judgments tied to specific institutional or evaluative settings.

The most studied examples are topic labelling and document classification. Kozlowski et al. evaluate LLM-based topic labelling by repeatedly applying FLAN, GPT-4, and GPT-4-mini to keyword representations generated by BERTopic (Kozlowski et al., 2024). They assess outputs using a combination of quantitative metrics and a qualitative accuracy coding scheme. Their results show that GPT-based models can generate labels that accurately describe topics derived from keyword sets.

Murray et al. formalise a distinction between descriptive and characteristic cluster labels and propose a structured workflow for generating and evaluating labels for clusters of scientific documents (Murray et al., 2025). In their demonstration, LLM-generated descriptive labels perform comparably to characteristic labels. The authors emphasise design choices and validation strategies as vital elements.

Related findings are reported for classification tasks more broadly. LLMs can approximate disciplinary similarity judgments at an aggregate level (Cantone et al., 2025), and LLM-assisted approaches can complement rule-based classification systems when applied to large publication sets (van Eck & Waltman, 2024). Across these studies, performance is more stable for broad categories (e.g. scientific fields or disciplines) than for fine-grained distinctions or individual documents.

GenAI performance is generally evaluated at an aggregate level, for example by comparing distributions of labels or classifications. Where validation against expert judgments is performed, results suggest that calibration and explicit validation procedures are necessary to assess reliability and to interpret disagreements (Murray et al., 2025). Prompting strategies can improve coherence and consistency but do not eliminate stochastic variation.

A second group of applications involves tasks where meanings can be inferred from text, but interpretation depends on contextual information that is not necessarily reflected in academic texts.

Examples include the inference of author characteristics (e.g. gender or academic rank), societal development goals (SDG) and reference function or sentiment analysis. A study retested SDG classification using LLMs and showed that performance varies across goals and depends strongly on prompt construction and context (Hajikhani & Cole, 2024). In a controlled evaluation of GPT-3.5 for citation context analysis, LLM-generated annotations were found to be as good or better than human annotations in terms of internal consistency but performed poorly in predictive performance when compared against human-labelled data (Nishikawa & Koshiba, 2024). The authors concluded that LLMs are not suitable as replacements for human annotators but may serve as auxiliary sources of information or as one annotator among several.

For author characteristics, a comparison of ChatGPT with established gender-inference APIs showed high accuracy (Goyanes et al., 2024). At the same time, limited reproducibility and sensitivity to contextual assumptions, such as nationality and cultural naming conventions, were documented.

A closely related example is author disambiguation, a longstanding problem in scientometrics that combines textual cues with relational information. Tuccari et al. showed that semantic information extracted from text using LLM improved performance when combined with co-authorship and citation indicators, whereas text-based inference alone was insufficient (Tuccari et al., 2025). The study illustrates how GenAI can contribute to tasks located at the boundary between semantic inference and contextual interpretation, provided that outputs are embedded in validated, hybrid workflows rather than used in isolation.

Consistent with our conceptual framework, techniques to provide GenAI with contextual information such as prompting increase the correlations between GenAI outputs and human annotations. Therefore, such outputs should be interpreted as probabilistic inference rather than definitive classifications, and validation against reference data remains necessary.

### 3.2 Pragmatics: evaluative and context-dependent judgments

The second group of applications of GenAI in scientometrics involve judgments that are meaningful only within specific settings, such as institutional or social contexts, and have an evaluative dimension. Examples are research evaluation, identifying prominent scholars, assessing novelty or disruptiveness, or predicting citation impact.

An evaluation of ChatGPT-4 on PLOS ONE abstracts identified three principal components from their rating criteria: quality and reliability, accessibility and understandability, and novelty and engagement (de Winter, 2024). The study finds that accessibility/understandability correlates with higher Mendeley readership, while they together with novelty/engagement relate to citation counts and social-media attention. In contrast, the quality/reliability component shows minimal correlation with impact measures. The results suggest that GenAI ratings primarily capture rhetorical and communicative features rather than evaluative judgments of scientific rigour.

Other research has shown that correlations between ChatGPT-generated evaluations and (human) peer review outcomes vary substantially across venues, with strong correlations in some settings and almost no correlation in others (Thelwall & Yaghi, 2025). ChatGPT fails to reliably identify prominent scholars at a single university, with results weakly correlated with citation counts or institutional status (Sandnes, 2024), while it can identify some landmark papers in astrophysics but also produces inconsistent judgments across repeated queries (Bornmann et al., 2024).

Several studies compared ratings of scientific articles by GenAI with self-assessments by authors (Thelwall, 2025b), with external expert ratings (de Almeida & Papanikolaou, 2025) and with internal departmental ratings (Langfeldt et al., 2026). The studies find that LLM-generated scores correlate with expert assessment, but also that correlations are higher when using only title and abstract as compared with full-text casting doubt on the depth of the evaluation. A large-scale study using UK Research Evaluation Framework scores as proxies of research quality showed moderate correlations with scores generated by ChatGPT, however with large variations by field (Thelwall, 2025a). From a scientometrics perspective, the most important result was that LLM-generated scores showed stronger correlations with human-generated scores than traditional bibliometric indicators.

### 3.3 Discussion

As expected from our framework, existing evidence suggested that GenAI performs relatively well on semantic inference tasks in scientometrics, particularly when outputs are used for descriptive or exploratory purposes rather than for definitive assignment (see Table 2 below).

In that respect, leveraging LLMs yields significant advancement compared to earlier natural language processing tools. The latter were often context-unaware and working entirely from the statistical properties of individual words or phrases with imposed monosemy. Thus, these kinds of tasks align closely with the strengths of LLMs.

Most studies, though, also point to the importance of calibration steps performed by human experts on a subsample of items to ensure robust and reliable performance. In keeping with these studies, we suggest that this should be considered a mandatory methodological practice for scientometric research using GenAI.

The studies we discussed in this section also showed variation associated with the level of context dependence of the meanings to be retrieved. GenAI performed better in applications internal to the academic context and with broad scope, such as labelling topics. Results were more mixed for applications with a lower level of granularity, such as analysing specialities, and when results could be interpreted only by relying on information from another context, such as the political discourse for SDGs.

Good practices suggested by the literature include using open-source (or open-weights) LLMs that can be (re-)trained (or fine-tuned) to the specific task at hand rather than generic-purpose GenAI tools. And it is recommended to rely on information from other sources, such as on institutional affiliation or nationality, to interpret results or to design the prompts.

Given the large and rapidly growing number of studies in this domain, we argue that the field now has to move from the current trial-and-error approach towards studies that focus on good practices in the design of GenAI applications. Specifically, applications should focus on semantics-oriented tasks where GenAI tends to perform well and simultaneously apply robust testing and validation techniques. This is particularly important for two core dimensions rarely addressed by current studies: the reproducibility and scalability of results.

As expected, the studies we discussed suggest that, while GenAI outputs in pragmatic tasks may show statistical regularities at an aggregate level, they are unstable and context-sensitive at the level of individual cases. This limits their usefulness for applications where judgments have direct, evaluative consequences. Several authors therefore conclude that GenAI may support exploratory analysis or summarisation but does not match the quality of human judgment in evaluative contexts (Thelwall, 2025c).

However, we suggest that GenAI could have a potential for large-scale evaluation on entire research fields to complement currently available bibliometric indicators. In that respect, there is some evidence that GenAI scores might mimic some aspects of academics' evaluation not fully covered by bibliometrics. Moreover, since large-scale evaluation reflect the overall collective judgments of academics in a field, rather than idiosyncratic individual evaluations, the probabilistic nature of LLMs might turn into an advantage.

We therefore argue that more in-depth theoretical and empirical research is needed on the specific mechanisms generating GenAI-based evaluation scores and on how they relate to the current bibliometric indicators in scientometrics.

*Table 2. GenAI task types and observed performance in scientometrics*

| Dominant textual function | Semantics | | Pragmatics | |
|---|---|---|---|---|
| Contextual dependence | Low (broad fields) | Medium (specialties, external context such as policy) | High (collective judgments) | Very high (individual judgments) |
| Typical tasks | Topic labelling, classification | Gender inference, reference function | Proxies of research quality, novelty at scale | Evaluation, prominence, novelty |
| Observed performance | Strong at aggregate level | Mixed, content-sensitive | Mixed, not fully understood | Limited |
| Main limitations | Stochastic variation | Cultural and temporal ambiguity | Lack of a clear theoretical framework | Context dependence, instability |
| Common mitigation strategies | Calibration, validation | Prompting, human review | Combination with bibliometric indicators | Human-machine integration |

# 4 GenAI influences on academia as an object of study

Thanks to its command of syntax and mastering of semantics, the currently most frequent use of GenAI in academia relates to writing or revising texts (Ding et al., 2025; Eger et al., 2025). This includes scholarly papers (Kobak et al., 2024), research proposals (Andersen et al., 2025), and reviews of papers and proposals (Zhou et al., 2024; Kousha & Thelwall, 2024).

While some of these uses are acknowledged by authors (Kousha, 2024), linguistic analyses suggested that explicit acknowledgments represent only a fraction of usage and that between one-quarter and one-third of scholarly papers in some fields are already written with at least some GenAI assistance (Kobak et al., 2024). A study showed that the presence of LLM-generated or -modified content has increased from approximately 2.5% to between 5% (in mathematics) and 20% (in computer science) on the sentence level, over the course of one year (Liang et al., 2024).

Yet, the textual content of scholarly publications represents the main empirical ground of scientometrics. This includes the words used (Leydesdorff, 1989), and authors and references listed (van Raan, 2004), which are considered as quantitative markers of the underlying cognitive and social processes in science (Wouters, 1999b; Leydesdorff, 2001). To this end, the field has long relied on stable distributions and almost natural law-like properties as indicators of cognitive, social, and institutional structures in science (Leydesdorff, 2001).

It is therefore highly relevant to understand whether large-scale generation of textual content by GenAI might alter the relationships between the social practices of science on the one hand, and the linguistic content of scholarly communication on the other hand. And whether this, in turn, requires a critical re-appraisal of the methods and standards of our field. In the following, we provide an explorative discussion of this issue, by focusing on four dimensions: changing vocabularies, authorship practices, productivity implications, and cited references.

## 4.1 Changing vocabularies

Recent studies provide evidence that texts generated with the help of GenAI are reshaping the linguistic features of scientific writing. This has primarily been measured and identified in abstracts. It is not clear if the findings can be generalized to full article texts.

Most straightforwardly this has been observed in a changed presence of specific words following the emergence of GenAI tools. For instance, a word such as *delve* had become a clear signal of AI-generated text but reversed growth after the discovery – either because people became aware, or newer models had updated frequencies (Leiter et al., 2024). This may indeed be a question of human awareness, as other highly consistent words (e.g. *is* and *are*) show significantly less frequent use in AI-generated texts and are also becoming less common in abstracts after the popularization of GenAI tools (Geng & Trotta, 2025).

Despite an increasing complexity and stylistic homogeneity, GenAI can also be seen as a linguistic equalizer, as the increasing complexity of text in abstracts can primarily be attributed to the writing of authors who are not native English-speakers (Lin et al., 2025). Consistently with these findings, explicit acknowledgments to ChatGPT for language editing primarily occurred in articles from non-English-speaking countries (Kousha, 2024). However, while LLM-generated texts have better syntactical quality than human-produced texts and are more complex (in terms of structure and vocabulary), they have also been found to be less readable with lower communicative relevance to their interlocutors (Fredrick & Craven, 2025; Zeleke et al., 2025; Alsudais, 2025). An analysis of 627,000 arXiv papers showed that younger, male, and non-native authors are also more likely to revise their work using GenAI (Zhu, W. & Cong, 2024). GPT-assisted revisions increase clarity, reduce hedging, and align writing with disciplinary norms favoured by senior researchers. While this may democratize participation, it also drives stylistic convergence and may reduce linguistic diversity in scientific communication.

Since co-word analysis is widely used in scientometrics to identify research topics and produce maps of science (Leydesdorff, 1989; Leydesdorff & Nerghes, 2017), we speculate that changes in vocabulary due to GenAI usage might affect these results. If text generation by GenAI aligns with disciplinary conventions, it might be expected that its use reinforces extant word associations and, accordingly, artificially increases the stability of research topics identified by scientometric techniques. This, in turn, might impair the identification of cognitive dynamics and of interdisciplinarity (Rafols & Meyer, 2010). At the same time, GenAI might introduce words in scientific texts based on associations that are not related to scientific content, and, therefore, lead to wrong conclusions on cognitive change.

Existing studies that investigate the linguistic impact of GenAI have focused on generic words that are of little interest for scientometrics. We therefore consider it urgent to replicate the same analyses on distinctive words associated with the cognitive content of scholarly work. We also suggest performing experimental studies where scientometric mapping techniques are applied to an original corpus of scientific papers and to the same corpus as revised by GenAI. Comparing the results in terms of word frequency, co-word association, and generated science maps should allow to isolate the effect of GenAI on these measurements.

### 4.2 Authorship practices

The integration of GenAI tools into scholarly writing raises questions for authorship norms, and particularly the responsibilities of disclosing tool usage (Formosa et al., 2024). The literature identifies a need for transparent, standardized practices for how GenAI should be acknowledged and cited (Hosseini et al., 2024). At the same time, there is no consensus around when AI “participation” should influence authorship attribution.

A bibliometric study of acknowledgments of and references to GenAI in academic publishing reveals a rising rate of adoption but inconsistent patterns (Gorraiz, 2025). The results are robust across disciplines, but with clearly broader adoption in Computer Science and Engineering. The study underscores ethical boundaries, which disallow AI as a co-author, due to its lack of intellectual responsibility. However, researchers have nonetheless experimented with ways to credit GenAI, triggering institutional responses and policy adjustments. A survey of members of the general public (holding at least a bachelor’s degree) suggests that perceptions of GenAI use in writing takes away from human credibility and requires full disclosure (Formosa et al., 2024). The same view, however, was also true for the use of human assistance, which was considered more deserving of credit. The authors, therefore, argue for a graded approach to authorship and disclosure, in which some types of GenAI contributions may qualify for authorship credit.

From a scientometric perspective, authorship is the most important operational proxy for intellectual contribution, collaboration, and responsibility. Even if GenAI is formally excluded from authorship, its systematic use in drafting, revising, or structuring texts may nonetheless alter the relationship between listed authors and the cognitive labour reflected in publications. This introduces a potential decoupling between authorship as recorded metadata and authorship as an indicator of contribution.

The heterogeneity of GenAI adoption across fields, countries and career stages implies that any effects on authorship practices are unlikely to be randomly distributed. Rather, GenAI may systematically affect observed collaboration patterns, for example by reducing the need for certain forms of linguistic, editorial, or statistical collaboration, or by lowering barriers to solo or small-team publication for specific groups of researchers.

For scientometrics, this raises empirical questions about the continued interpretability of authorship-based indicators, such as co-authorship rates, team sizes, and measure of collaboration intensity. We strongly recommend that this should be monitored through empirical analyses of co-authorship patterns over time, combined with analyses of AI generated content through linguistic and semantic markers. Monitoring longitudinal changes in the indicators could help distinguish structural shifts in collaboration from changes driven by altered writing practices.

## 4.3 Affecting productivity measures

Productivity is a central construct in scientometrics, commonly operationalised through publication counts. These measures implicitly assume relatively stable relationships between effort, output, and contribution, as well as broadly comparable production technologies across researchers.

Recent evidence suggests that GenAI use may alter these relationships since researchers using GenAI tools publish more frequently, but also with a tendency toward increased concentration on established research topics (Hao et al., 2024). Accordingly, it has also been reported that productivity gains may coincide with reduced topical exploration (Kusumegi et al., 2025).

These findings raise questions about the interpretation of productivity indicators. If GenAI systematically lowers the marginal cost of producing publishable results and academic texts, increases in output may no longer be comparable across time or groups with different access to or adoption of these tools. Moreover, if productivity gains are coupled with reduced topical diversity, traditional indicators may misinterpret scientific progress or innovation.

Such developments may also interact with well-established scaling properties of scientific production, including cumulative advantage processes and highly skewed productivity distributions (Ruiz-Castillo & Costas, 2014). Empirical studies show that citation distributions are not static but have become increasingly concentrated over time (Nielsen & Andersen, 2021). This indicates that shifts in production or dissemination practices can affect not only average output but the structure of scientific inequality. Whether recent changes in writing and production practices amplify or dampen existing disparities remains an open empirical question.

## 4.4 Use and meaning of references

Shortly after the release of ChatGPT, research showed that the tool was inclined to fabricate or *hallucinate* scholarly references (Orduña-Malea & Cabezas-Clavijo, 2023). This finding has been confirmed in several other studies, though more recently substantial improvement has been shown in the degree to which this happens (Walters & Wilder, 2023; Lehr et al., 2024).

While reference fabrication is an issue for science, it is less of a problem for scientometrics. Typically, citation studies focus on the cited object (which is real), but in the case of reference-focused analyses, fabrication may play a role as a source for lower data quality.

More importantly though, the use of generative AI tools may potentially change the sources that scientists discover, if such tools are used in the place of systematic literature searches. GenAI-generated references might reinforce trends already generated by tools such as Google Scholar for the identification and semi-automated citation of sources (Drivas, 2025). In our view, this might require revisiting the theoretical foundations for citation analysis (Wouters, 1999a) and re-calibrating widely used methods and indicators.

First, commonly found citation patterns based on preferential attachment, social ties, status etc. might be affected. Initial results suggest that reference selection of GenAI tools such as ChatGPT affirm human biases to prefer highly cited research (Algaba, Mazijn et al., 2025; Algaba, Holst et al., 2025). The impacts might be at different levels: on the one hand, social network maps of science based on (co)-citations might become more hierarchical and centred on a few stars. On the other hand, the interpretation of reputational indicators (at the individual and institutional level) might become problematic as the underlying scaling laws are affected by GenAI (Peterson et al., 2010).

Second, if references were generated automatically by GenAI, the long-standing assumption in science that cited references can be used to trace intellectual traditions and diffusion of novel ideas becomes less defendable (Uzzi et al., 2013). While issues with the meaning of citations and the underlying processes are well-known (Tahamtan & Bornmann, 2018), GenAI could effectively reverse the direction of the process, i.e. inserting references based on word association with *existing* textual content rather than deriving new ideas from references.

While large-scale impacts will be difficult to observe and (causally) connect to GenAI, we suggest a more limited empirical protocol where references are stripped from a corpus of scientific papers and then re-generated using GenAI – and the structure of the two sets of references is then systematically compared. Empirical studies of the extent scholars are adopting GenAI for the discovery of references to be cited would be also as important.

## 4.5 Discussion

The arguments presented in the previous sections suggest that, indeed, GenAI has the potential to alter some of the key measures used in scientometrics. Given that the relationships between quantitative measures derived from textual content of publications and scientific practices represents the core conceptual basis of the field (Wouters, 1999b; Leydesdorff, 2001), we consider that this issue cannot be further neglected by our community.

We also propose two directions for research. First, there is a need to revisit the theoretical and methodological roots of scientometrics measures, such as using citations as a proxy for social structure of science. We also have to inquire whether text generation by GenAI is altering some basic assumptions on scientific practices on which these measures are rooted. Second, the rapid diffusion of GenAI in academic writing constitutes a natural experiment, which would allow for a comparison before and after GenAI of the distribution of key measures, such as scientific keywords, co-authorship and citation patterns. This would provide some empirical evidence of whether the effects we hypothesized are taking place.

# 5 Discussion and conclusions

In this opinion paper, we addressed two questions, i.e. a) can we propose a framework to interpret the performance of GenAI in scientometrics, as related to the nature of the task, and to devise robust adoption strategies and b) can we understand the implications of the (automated) generation of large amounts of scientific language by GenAI on core measures in scientometrics, such as research topics, authorships, productivity and references?

Given the novelty and importance of the topic for the scientometrics community, our aim was to provide conceptual elements, supported by the so far still rather limited available empirical evidence. And we used this, in turn, to identify priority areas for further research.

On the first question, we have argued that GenAI tools are expected to perform systematically better for semantic tasks, i.e. retrieving the meanings attributed to words, than for pragmatic tasks, i.e. generating content to become actionable in specific communicative situations. While we should be cautious in drawing conclusions on what GenAI can and cannot do given the rapid advances in LLM performance, these patterns are associated with fundamental design features of current tools and, accordingly, are unlikely to fundamentally change within the current generation of models.

Our analysis of GenAI's scientometrics applications broadly confirmed these expectations. LLMs perform relatively better on tasks related to language generation, such as topic labelling and summarization. The performance is weaker when tasks require stability and reproducibility such as the inference of author characteristics or classifying the function of scholarly references. In this area, we put forward the suggestion that scientometrics research should now move towards methodological consolidation and developing good practices for testing and validation, especially with a view towards the reproducibility of results.

In contrast, when dealing with evaluative applications, such as rating papers, peer reviewing papers or research evaluation GenAI outputs, may show statistical regularities at an aggregate level but they are unstable at the level of individual cases. This limits their usefulness for applications where judgments have direct, evaluative consequences. However, we suggested that GenAI might complement current bibliometric indicators for large-scale evaluations, an area which would clearly deserve more in-depth investigation.

In our view, these results have implications for the application of LLMs to scientometrics, and the use of the results they generate. First, and most importantly, scholars should identify the nature of the tasks assigned to LLMs – is it to summarize an existing corpus of knowledge, or to detect statistical regularities, or to identify and/or evaluate individual scholars. Second, the level of granularity of the analysis represents a key parameter – is it the goal to detect large-scale relationships, or to identify statistical associations by broad contexts such as scientific fields, or to provide reliable pointwise estimates for a single paper or an individual scholar. And third, the purpose of the analysis matters – is it to describe patterns and detect regularities, or the results are an input for practical tasks such as evaluation and distribution of funding.

These task characteristics should, first, inform the decision on whether to rely on GenAI or not. And second, they should allow scholars to devise strategies for human-machine interaction – when detecting large-scale regularities, the role of humans might be focused on calibration and validation of results, while in evaluative tasks GenAI could function rather as an assistant for human evaluation.

On the second question, our arguments were largely speculative as there is virtually no research yet on the impact of GenAI on scientometrics. However, our reasoning suggests that GenAI has the potential to alter some core measures used in the field, such as vocabulary, acknowledgement and authorship practices, the number of papers produced and the way references are generated and used. We remarked that these are core quantitative markers of the underlying cognitive and social processes in science. Accordingly, we suggested that it becomes urgent to undertake empirical studies of implications for scientometrics measures, such as analysing changes in the scientific vocabulary and in the use of references in papers that might be associated with GenAI use.

In conclusion, GenAI is not just a new method or productivity tool. It has the potential to alter the textual representation of science, which can pose both a challenge and opportunity for scientometrics. It challenges the reliability of established indicators, but it might also open new avenues for modelling and studying scientific communication. Careful empirical work and theoretical reflection will be essential to ensure the field remains capable of interpreting and explaining the evolving patterns of knowledge production in the age of AI.

# 6 Acknowledgements

The authors acknowledge support and constructive comments from the journal editorial team and from three anonymous reviewers, which helped us immensely to develop our arguments. Many of the ideas presented were first discussed at the Scientific workshop on using generative AI for science and higher education studies, which took place at the Università della Svizzera Italiana in Lugano in February 2025 and was supported by the Swiss National Science Foundation.

# 7 Funding

This work is supported by the Swiss National Science Foundation (Grant No. 230119).